\pdfoutput=1

\documentclass[11pt]{article}

\usepackage[final]{acl}

\usepackage{times}
\usepackage{latexsym}
\usepackage{multirow}
\usepackage{booktabs}
\usepackage{makecell}
\usepackage{amssymb}
\usepackage{enumitem}
\usepackage[T1]{fontenc}

\usepackage[utf8]{inputenc}

\usepackage{microtype}

\usepackage{inconsolata}
\usepackage{CJKutf8}

\usepackage{graphicx}
\usepackage{amsmath}
\usepackage{amsfonts}
\usepackage{colortbl}
\usepackage{fontawesome5}
\usepackage{float}

\usepackage{tikz}

\title{$\mathcal{V}isi\mathcal{P}runer$: Decoding Discontinuous Cross-Modal Dynamics for \\ Efficient Multimodal LLMs}

\author{
 \textbf{Yingqi Fan\textsuperscript{1}},
 \textbf{Anhao Zhao\textsuperscript{1,3}},
 \textbf{Jinlan Fu\textsuperscript{5}},
 \textbf{Junlong Tong\textsuperscript{1,2}},
\\
 \textbf{Hui Su\textsuperscript{4}},
 \textbf{Yijie Pan\textsuperscript{1}},
 \textbf{Wei Zhang\textsuperscript{1}},
 \textbf{Xiaoyu Shen\textsuperscript{1}\thanks{Corresponding Author}}
\\
 \textsuperscript{1}Ningbo Key Laboratory of Spatial Intelligence and Digital Derivative,
\\
Institute of Digital Twin, EIT, Ningbo
 \textsuperscript{2}Shanghai Jiao Tong University
\\
 \textsuperscript{3}Hong Kong Polytechnic University
 \textsuperscript{4}Meituan Inc.
 \textsuperscript{5}National University of Singapore
\\
 \small{
   \textbf{Correspondence:} \href{mailto:email@domain}{yingqi949@gmail.com} \href{mailto:email@domain}{xyshen@eitech.edu.cn}
 }
}

\begin{document}
\maketitle
\begin{abstract}
Multimodal Large Language Models (MLLMs) have achieved strong performance across vision-language tasks, but suffer from significant computational overhead due to the quadratic growth of attention computations with the number of multimodal tokens. Though efforts have been made to prune tokens in MLLMs, \textit{they lack a fundamental understanding of how MLLMs process and fuse multimodal information.} Through systematic analysis, we uncover a \textbf{three-stage} cross-modal interaction process: (1) Shallow layers recognize task intent, with visual tokens acting as passive attention sinks; (2) Cross-modal fusion occurs abruptly in middle layers, driven by a few critical visual tokens; (3) Deep layers discard vision tokens, focusing solely on linguistic refinement. Based on these findings, we propose \emph{VisiPruner}, a training-free pruning framework that reduces up to 99\% of vision-related attention computations and 53.9\% of FLOPs on LLaVA-v1.5 7B. It significantly outperforms existing token pruning methods and generalizes across diverse MLLMs. Beyond pruning, our insights further provide actionable guidelines for training efficient MLLMs by aligning model architecture with its intrinsic layer-wise processing dynamics. Our code is available at: \url{https://github.com/EIT-NLP/VisiPruner}.
\end{abstract}

\definecolor{lightGray}{RGB}{186, 186, 186}
\definecolor{customGreen}{RGB}{84, 158, 63}
\definecolor{customRed}{RGB}{222, 93, 78}

\section{Introduction}

Multimodal Large Language Models (MLLMs) \citep{multimodal_llm_survey} extend the reasoning power of Large Language Models (LLMs) to other modalities like vision \citep{li2023blip2bootstrappinglanguageimagepretraining}, audio \citep{guzhov2021audioclipextendingclipimage}, and video \citep{alayrac2022flamingovisuallanguagemodel,tong2025context}, typically by aligning modality encoders (e.g., ViT \citep{vit}) with LLMs through lightweight projectors \citep{liu2023llava,lin2025multilayervisualfeaturefusion,chen2025rethinkingvisuallayerselection,zhao2024unveilingincontextlearningcoordinate}.
However, visual encoders often produce far more tokens than text due to higher information density. This not only inflates the sequence length but also results in a quadratic increase in attention computation. While recent efforts like token pruning \citep{fit_and_prune,llava_prumerge,lin2024preservecompressindepthstudy}, dynamic resolution \citep{attention_guided_token_dropping,monkey}, and sparse attention mechanisms \citep{treat_visual_tokens_as_text?,beyond_token_compression} aim to mitigate this issue, their effectiveness remains limited due to \emph{a fundamental gap in understanding how MLLMs actually process and integrate visual information across layers.}

\begin{figure}[t]
    \centering
    \includegraphics[width=1\linewidth]{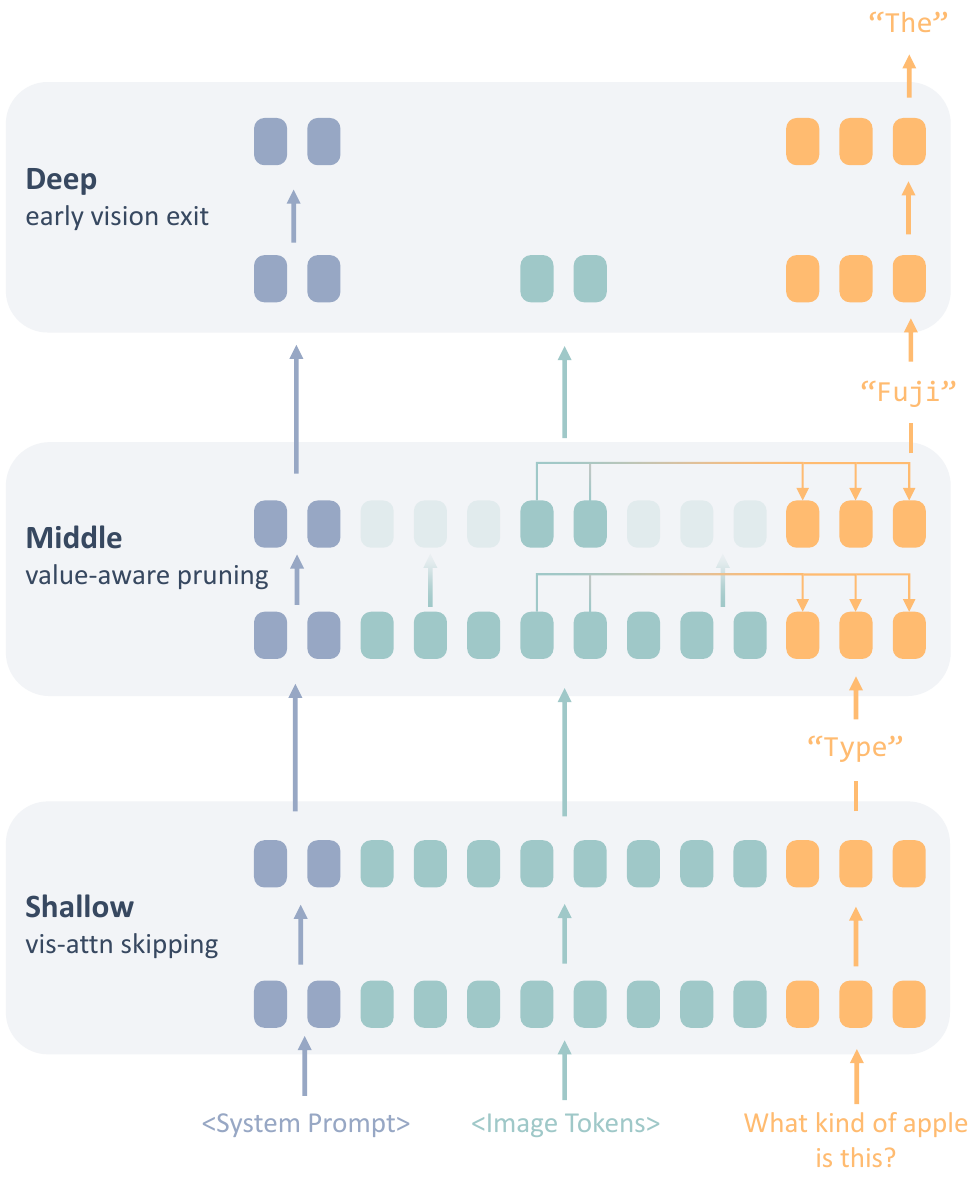}
    \vspace{-25pt}
    \caption{\small \textbf{Illustration of the three-stage discontinuous information processing in Multimodal Large Language Models (MLLMs)}. The framework separates visual-text integration into three key stages: \textbf{Shallow Layers} focus on task recognition, \textbf{Middle Layers} highlight the cross-modal fusion of sparse, task-relevant visual tokens, and \textbf{Deep Layers} focus on linguistic refinement after vision integration.}
    \label{fig:overall_framework}
\end{figure}

Existing analyses of cross-modal interactions in MLLMs predominantly rely on attention scores as proxies for information flow \cite{wu2024acceleratingmultimodallargelanguage, zhang2025llavaminiefficientimagevideo, zhang2024crossmodalinformationflowmultimodal}. This has led to widespread but misleading conclusions, e.g., the assumption that cross-modal fusion mainly occur in shallow layers. We  move beyond attention maps to understand how and when visual information is actually utilized, revealing three insights that revise the current understanding of MLLMs:
\begin{itemize}[leftmargin=*, labelsep=0.5em]
    \item \textbf{Shallow Layers as Task Recognizers}: Contrary to prior beliefs ~\cite{wu2024acceleratingmultimodallargelanguage,zhang2025llavaminiefficientimagevideo,zhang2024crossmodalinformationflowmultimodal}, cross-attention in early layers serves \emph{no meaningful role} in visual-text fusion. Visual and textual tokens evolve \emph{independently}, with shallow layers functioning solely to recognize task from text instructions, while visual tokens act merely as `attention sinks''~\cite{xiao2024efficientstreaminglanguagemodels}.
    \item \textbf{Sparse Critical Tokens in Middle Layers}: Cross-modal integration occurs abruptly in intermediate layers, but only \emph{a few critical visual tokens drive this process}. Conventional attention-based methods fail to identify these tokens, as their importance correlates with feature similarity rather than attention weights.
    \item \textbf{Instruction Alignment in Deep Layers}: Once visual information has been integrated into the text encoder, deeper layers \emph{discard vision tokens} and transition to pure linguistic refinement to output final answers.
\end{itemize}
Building on these insights, we introduce \emph{VisiPruner}, a training-free pruning framework that exploits both layer-wise and token-wise redundancy in MLLMs. For layer-wise compression, our method disables cross- and self-attention in shallow visual layers and removes visual tokens in deep layers, allowing seamless integration with existing token pruning methods. For token-wise compression, we propose a novel influence-based method to dynamically identify and retain only the most interactive visual tokens from middle layers. Together, these strategies reduce up to 99.0\% of visual-related attention computations and 53.9\% of total FLOPs, all while preserving performance across a range of MLLMs and benchmarks.

Our findings further offer actionable guidelines for designing efficient MLLMs.
While \emph{VisiPruner} demonstrates the principles in a training-free paradigm, embedding them directly into MLLM training pipelines should further optimize performance-efficiency tradeoffs. 
Overall, our work makes four key contributions: (1) To the best of our knowledge, we are the first systematic analysis revealing the discontinuous, sparse, and decoupled nature of cross-modal interactions in MLLMs, particularly highlighting the counter-intuitive finding that \emph{shallow layers operate independently of vision}; (2) Exposing the inadequacy of attention-based analysis for understanding visual token utility by attention merging; (3) A training-free pruning framework validated across diverse MLLMs and benchmarks; and (4) Actionable guidelines for designing efficient MLLMs that align with their intrinsic mechanics.

\section{Background}
\label{sec:bg}
Modern MLLMs integrate perceptual modalities (e.g., vision) with linguistic reasoning using a vision encoder, projection, and language backbone~\cite{llava, chu2024mobilevlmv2fasterstronger}.

\paragraph{Modality-Specific Encoding}
Let input \( v \in \mathcal{V} \) (e.g., an image) and text instruction \( x \in \mathcal{X} \). Each modality is encoded independently:
\vspace{-6pt}\[
\begin{aligned}
\text{Visual encoder:} \quad & \mathbf{E}_v = \mathcal{V}(v) \in \mathbb{R}^{N_v \times d_v}, \\
\text{Textual encoder:} \quad & \mathbf{E}_t = \mathcal{T}(t) \in \mathbb{R}^{N_x \times d_x},
\end{aligned}
\]

where \( \mathcal{V} \) (e.g., ViT) and \( \mathcal{T} \) (i.e., LLM tokenizer) map inputs to sequences of embeddings. Typically, \( N_v \gg N_x \) due to the high information density of \( \mathcal{V} \), e.g., \( N_v = 576 \) for a \( 336 \times 336 \) image with patch size \( 14 \)~\cite{chen2024imageworth12tokens}.

\paragraph{Cross-Modal Projection}
A projector \( \mathcal{P} \) aligns visual embeddings to the LLM’s text space $\mathbf{H}_x^{(0)}$:
\vspace{-5pt}\[
\mathbf{H}_v^{(0)} = \mathcal{P}(\mathbf{E}_v) \in \mathbb{R}^{N_v \times d_h},
\]\vspace{-5pt}
where \( d_h \) matches the LLM’s hidden dimension. 

\paragraph{Layer-Wise Cross-Modal Fusion}
The fused input is defined as \( \mathbf{H}^{(0)} = \mathbf{H}_v^{(0)} \oplus \mathbf{H}_t^{(0)} \) (\( \oplus \) denotes concatenation), which is processed through \( L \) transformer layers. At layer \( l \), cross-attention and self-attention are computed as~\cite{zhao2025skipgptdynamiclayerpruning}:
\[
\mathbf{Q}_t^{(l)} = \mathbf{H}_t^{(l-1)}\mathbf{W}_Q, \quad 
\mathbf{K}_v^{(l)}, \mathbf{V}_v^{(l)} = \mathbf{H}_v^{(l-1)}\mathbf{W}_{K/V},
\]
\[
\mathbf{A}^{(l)} = \text{softmax}\left(\frac{\mathbf{Q}_t^{(l)}(\mathbf{K}_v^{(l)})^\top}{\sqrt{d_h}}\right) \in \mathbb{R}^{N_x \times N_v},
\]
\[
\mathbf{H}_{\text{cross}}^{(l)} = \mathbf{A}^{(l)}\mathbf{V}_v^{(l)},
\]
\[
\mathbf{H}_t^{(l)} = \text{TransformerBlock}(\mathbf{H}_t^{(l-1)} + \mathbf{H}_{\text{cross}}^{(l)}),
\]

A key computational bottleneck in MLLMs arises from the large number of visual tokens~\cite{zhang2025llavaminiefficientimagevideo}. In most scenarios, \( N_v \gg N_x \), the cross-attention matrix \( \mathbf{A}^{(l)} \in \mathbb{R}^{N_x \times N_v} \) grows significantly, making its computation a dominant cost factor. We believe that not all visual tokens contribute meaningfully to text-driven reasoning. To address this, we seek to (1) \emph{Understand cross-modal information flow} and (2) \emph{Reduce unnecessary visual-text interaction}.
\section{Shallow Layers: Task Recognition}
\label{sec:shallow}

Shallow layers in MLLMs are often assumed to be crucial for cross-modal fusion due to two observations: (1) High cross-attention scores between instruction tokens and vision tokens in early layers~\cite{wu2024acceleratingmultimodallargelanguage,zhang2025llavaminiefficientimagevideo}; and (2)  Performance degradation when cross-attention in shallow layers is masked~\cite{zhang2024crossmodalinformationflowmultimodal}.
We systematically re-evaluate these claims and present evidence that contradicts these assumptions.

\subsection{Attention Scores $\neq$ Information Utility}
Although attention scores are often interpreted as measures of token importance, we provide two key counterarguments that challenge this assumption.
\paragraph{Counterpoint 1: Static Attention Patterns} We first visualize attention maps across shallow, middle and deep layers (\autoref{app:vis_sink_token}). A striking pattern emerges: the most attended vision tokens remain unchanged regardless of the input instruction in shallow layers. Whether the task involves color identification (e.g., “What color is the dog?”) or scene understanding (e.g., “Is there any scooter?”), the same image regions consistently receive the highest attention. Although it is counterintuitive that different tasks attend to the same visual features, these visual tokens may contribute to global understanding of the image \cite{darcet2024visiontransformersneedregisters}.

\paragraph{Counterpoint 2: Masking Highly Attended Tokens has No Effects}To further test if highly attended tokens encode global information, we mask the top 10\% most attended vision tokens in layers 1–2 and evaluate performance. If these tokens were essential, their removal should degrade performance. However, results show minimal change (\autoref{tab0:masking_top_tokens_in_shallow_layers}).
This directly contradicts the claim that attention scores reflect information utility. Apparently, \emph{high attention scores in shallow layers do not imply high information utility}. 
\begin{table}[h!]
\centering \footnotesize
    \renewcommand{\arraystretch}{1.1}
    \begin{tabular}{@{\hskip 2pt}l@{\hskip 6pt}ccccc@{\hskip 2pt}}
    \toprule
        \textbf{Model} & \textbf{GQA} & \textbf{MME$^{P}$} & \textbf{POPE}  & \textbf{MMB} \\
    \midrule
        \cellcolor{gray!20} LLaVA-v1.5 7B   & \cellcolor{gray!20}62.0  & \cellcolor{gray!20}1507.6 & \cellcolor{gray!20}85.9  & \cellcolor{gray!20}64.3\\
        + Mask  & 62.0 & 1506.6 & 85.7  & 64.3 \\
    \midrule
        \cellcolor{gray!20}LLaVA-v1.5 13B   & \cellcolor{gray!20}63.3 & \cellcolor{gray!20}1531.3 & \cellcolor{gray!20}85.9  & \cellcolor{gray!20}67.7\\
        + Mask  & 63.2 & 1518.6 & 86.3  & 68.9 \\
    \midrule
        \cellcolor{gray!20}InternVL2.5 8B   & \cellcolor{gray!20}63.6 & \cellcolor{gray!20}1700.0 & \cellcolor{gray!20}90.6  & \cellcolor{gray!20}84.6\\
        + Mask  & 63.2 & 1689.5 & 90.6  & 84.3\\
    \midrule
        \cellcolor{gray!20}MobileVLM-v2 3B   & \cellcolor{gray!20}61.0 & \cellcolor{gray!20}1440.5 & \cellcolor{gray!20}84.7  & \cellcolor{gray!20}63.2\\
        + Mask  & 60.9 & 1440.8 & 84.6  & 63.3\\
    \bottomrule
    \end{tabular}
\vspace{-5pt}
\caption{Performance after masking top 10\% attended visual tokens in the first two layers on diverse MLLMs. See~\autoref{app:shallow-attended_masking} for results under different selection criteria.}
\label{tab0:masking_top_tokens_in_shallow_layers}
\end{table}

\subsection{Redundant but Necessary?}
\label{subsec:shallow_halfMasking_kvRemoval}
Given that high attention $\neq$ information utility, we now examine whether shallow-layer visual tokens serve any information utility at all.
\paragraph{The Redundancy Paradox}  Following our previous result that masking top 10\% most attended  tokens has no effect, if cross-modal fusion does occur in shallow layers, it would have to reside in the remaining 90\% of tokens. We now mask the remaining 90\% tokens to see if these tokens alone are sufficient for multimodal fusion. Surprisingly, we again find minor degradation in overall accuracy\footnote{Averaged over four benchmarks (GQA, MME, POPE and MMB) and two MLLMs (LLaVA-v1.5 7B and 13B). Note that the score for MME is divided by 20 before averaging.} (\( 72.6 \to 71.5 \), suggesting that \emph{neither the most attended nor the least attended vision tokens carry essential information!} To further probe the necessity of individual tokens, we randomly mask half of the visual tokens and measure performance changes:

\begin{itemize}[leftmargin=*, labelsep=0.5em, itemsep=0pt]
    \item \textbf{Left Half Masking} (removing first 288 of 576 tokens): \( 72.6 \to 72.6 \)
    \item \textbf{Right Half Masking} (removing last 288 tokens): \( 72.6 \to 72.4 \).
\end{itemize}

We can see that the performance remains stable regardless of which tokens are masked, implying that \emph{visual tokens in shallow layers are largely redundant in terms of content transfer}. 

During decoding, we observe the same as in pre-filling stage (see \autoref{app:decoding}), confirming the absence of cross-modal fusion in shallow layers. It is likely that visual tokens in shallow layers do not contribute to information fusion in a meaningful way. Instead, their presence—regardless of which specific tokens remain—appears to be necessary for stability rather than content transfer.

\subsection{Vision as Attention Stabilizers}
\label{subsec:shallow_merging}

\begin{figure}[t!]
    \centering
    \includegraphics[width=0.8\linewidth]{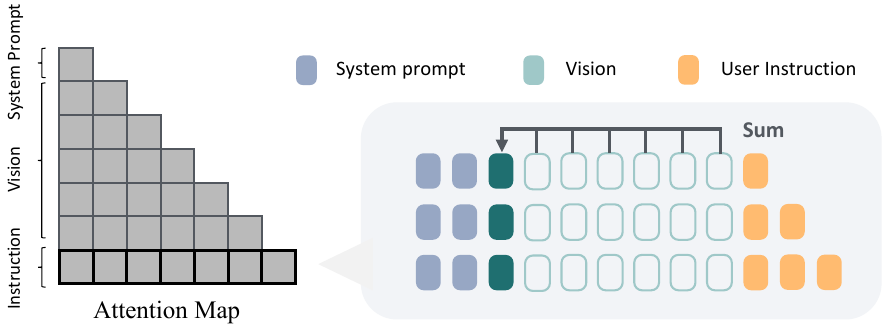}
    \vspace{-10pt}
    \caption{\small Merge visual attention weights into a single token to stabilize the attention distribution of the first layer.}
    \label{fig:layer0_masking}
\end{figure}
Given that no individual vision tokens carry essential cross-modal information, it is paradoxical that cross attention knockout in shallow layers leads to significant loss in visual perception \cite{zhang2024crossmodalinformationflowmultimodal,geva2023dissectingrecallfactualassociations}. Hence, we hypothesize that their role is to stabilize shallow-layer attention distributions without transmitting meaningful content. To test this, we propose \emph{Attention Merging}, forcing all cross-attention weights in shallow layers to focus on a single visual token (\autoref{fig:layer0_masking}):
\vspace{-6pt}\begin{equation}
    \mathbf{A}^{(l)}_{i,j} = \begin{cases}
            \sum_{v\in V}A^{(l)}_{i,v} & \text{if } j = k \\
            0 & \text{otherwise}
        \end{cases}
\end{equation}
where \( V \) represents all vision tokens and \( k \) is the randomly selected index of the merged token. If shallow vision tokens were performing useful fusion, constraining attention to a single token should degrade performance. However, we observe no meaningful change across different choices of \( k \) (see ~\autoref{app:random_merging}), confirming that \emph{no specific vision token is necessary for shallow-layer computation}. The model simply requires \emph{some} tokens to absorb attention weights. Even further, we show that the stabilization is needed only in the first layer:

\begin{itemize}[leftmargin=*, labelsep=0.5em, itemsep=0pt]
    \item \textbf{layer 1}:  masking all vision tokens significantly degrades average performance (\( 72.6 \to 65.2 \)), confirming that a visual attention sink is needed.
    \item \textbf{layer 2-7}:  system prompts can replace vision tokens as attention sinks, with no performance drop (\( 72.6 \to 72.1 \)) (see \autoref{app:analysis_vis_sinks}). 
\end{itemize}
This dichotomy arises from diverging value vector distributions: early vision token values ($\mathbf{V}_v^{(0)}$) differ significantly from text tokens ($\mathbf{V}_x^{(0)}$), necessitating modality-specific sinks initially (\autoref{app:l1_norm_value}).

Overall, these results suggest that \emph{shallow-layer vision tokens primarily serve as a stabilization mechanism for attention, rather than contributing to meaningful cross-modal fusion of information}.

\subsection{Null Effects in Decoding Stage}
\label{app:decoding}

The cross-modal fusion happens in two stages: 
\begin{itemize}[leftmargin=*, labelsep=0.5em, , itemsep=0pt]
    \item \textbf{Prefill Phase}: The entire input sequence, including visual and text embeddings, is processed in a single forward pass. This initializes hidden states for subsequent decoding.
    \item \textbf{Decoding Phase}: Tokens are generated autoregressively, where each new token attends to previously generated tokens while interacting with visual representations.
\end{itemize}

Apart from the prefilling stage, we also remove vision tokens from the key-value (KV) cache at different depths in the decoding stage. As can be seen in \autoref{tab2:mmvet}, the result is even better after removing the KV cache (see \autoref{app:mmvet_breakdown}). This further supports our claim that shallow visual tokens do not meaningfully contribute to content information. Instead, their role appears to be largely structural rather than informational.

\begin{table}[h!]
\centering \footnotesize
    \begin{tabular}{cc|cc}
    \toprule
        \textbf{Model} & \textbf{Layers} & \textbf{MM-Vet} & \textbf{GQA} \\
   \midrule
        \multirow{5}{*}{\textbf{LLaVA-v1.5 7B}} & \cellcolor{lightGray!30}-       & \cellcolor{lightGray!30}31.2 & \cellcolor{lightGray!30}62.0 \\
           & 1--8    & 33.8 & 61.8 \\
           & 9--15   & 28.3 & 61.8 \\
           & 26--32  & 31.1 & 61.9 \\
           & 1--32   & 26.1 & 61.7 \\
        \bottomrule
    \end{tabular}
\caption{\label{tab2:mmvet} \textbf{Performance with visual information removed from specific KV Cache layers.} MM-Vet is a benchmark requiring key visual information to remain in the KV Cache \cite{benchmark:mmvet}.}
\end{table}

\subsection{Role of Shallow Layers}
\label{sec:shallow_role}
Having confirmed the absence of meaningful cross-modal information flow, and the visual and text layers evolve largely independently. We further investigate the actual roles of shallow layers.

\paragraph{Shallow Text Layers: Task Recognition} To analyze what shallow text layers are mainly doing, we analyze the semantic content of the final token’s hidden state by projecting it through the model's unembedding matrix~\cite{interpreting_gpt}:
\vspace{-5.5pt}
\begin{equation}
    D_{\text{last}} = \text{softmax}(W_u h_{\text{last}}^\ell),
    \label{eq:last_token_projection}
\end{equation}
where \( D_{\text{last}} \) represents the probability distribution over vocabulary tokens. We find that shallow-layer representations align with task semantics rather than visual content.  
For example, intermediate layers produce activations aligned with task-relevant words:  
- ``\textit{How many cars...}'' → ``number'' (Layer 10)  
- ``\textit{What kind of...}'' → ``type'' (Layer 7) (\autoref{app:projection}). 

Beyond the latent representation of the final input token, we further observe that the value-output matrix also encodes task information in shallow layers, reinforcing our finding (\autoref{app:proj_vo}).
\begin{equation}
    D_{\text{vo}} = \text{softmax}(W_u \cdot V_{\text{last}}^\ell \cdot O),
    \label{eq:vo_projection}
\end{equation}

These findings suggest that \emph{shallow text layers are primarily responsible for task recognition, operating independently from visual processing}.

\paragraph{Shallow Visual Layers: Feature Alignment}  Knowing that little cross-modal interaction is performed in shallow visual layers, we further investigate whether intra-modal fusion occurs. Specifically, we mask self-attention among visual tokens, forcing each token to be processed independently. As shown in \autoref{tab1:masking_shallow_layers}, this modification results in only a minimal performance drop, indicating that self-attention plays a negligible role.

These results suggest that \emph{the primary function of shallow visual layers is neither cross, nor intra-modal fusion, but rather the alignment of ViT features with the LLM’s internal representation space, implying that the attention mechanism in these layers may be largely redundant}.

\begin{table}[h!]
\centering \footnotesize
    \begin{tabular}{cccc  c}
    \toprule
        \textbf{Layers} & \textbf{Masking} & \textbf{\#Token}  & \textbf{Merging} & \textbf{GQA}\\
    \cmidrule(lr){1-4}\cmidrule(lr){5-5}
        \cellcolor{lightGray!30} -   & \cellcolor{lightGray!30}-   & \cellcolor{lightGray!30}N/A   & \cellcolor{lightGray!30}N/A  & \cellcolor{lightGray!30}61.95 \\
        \multirow{4}{*}{1-2} 
            & \multirow{2}{*}{C}    
                  & 576 & No  & 57.41 \\
                & & 575 & Yes & \textbf{61.98} \\
            \cmidrule(lr){2-4}\cmidrule(lr){5-5}
            & \multirow{2}{*}{C\&V} 
                  & 576 & No  & 56.08 \\
                & & 575 & Yes & \textbf{61.96} \\
    \cmidrule(lr){1-4}\cmidrule(lr){5-5}
        \multirow{4}{*}{1-7} 
            & \multirow{2}{*}{C}    
                  & 576 & No  & 57.18 \\
                & & 575 & Yes & \textbf{61.51} \\
            \cmidrule(lr){2-4}\cmidrule(lr){5-5}
            & \multirow{2}{*}{C\&V} 
                  & 576 & No  & 54.63 \\
                & & 575 & Yes & \textbf{60.78} \\
    
    \hline
    \end{tabular}
\vspace{-5pt}
\caption{\small \textbf{Impact of vision on cross-attention stability.} \textit{Layers} refer to layers with attention masked. \textit{\# Tokens} indicates the number of masked vision tokens. ``C'' represents cross attention masking; ``V'' represents visual self attention masking. }
\label{tab1:masking_shallow_layers}
\end{table}

\subsection{Strategy for Efficient MLLM Design}
Based on these findings, we propose a simple yet effective pruning strategy for shallow layers: 
(1) Merge visual attention in layer 1 to serve as an attention sink; (2) Skip visual-textual attention computation for all vision tokens in layers 2+; and (3) Remove visual self-attention.  

\begin{figure}[t!]
    \centering
    \includegraphics[width=1\linewidth,height=5cm]{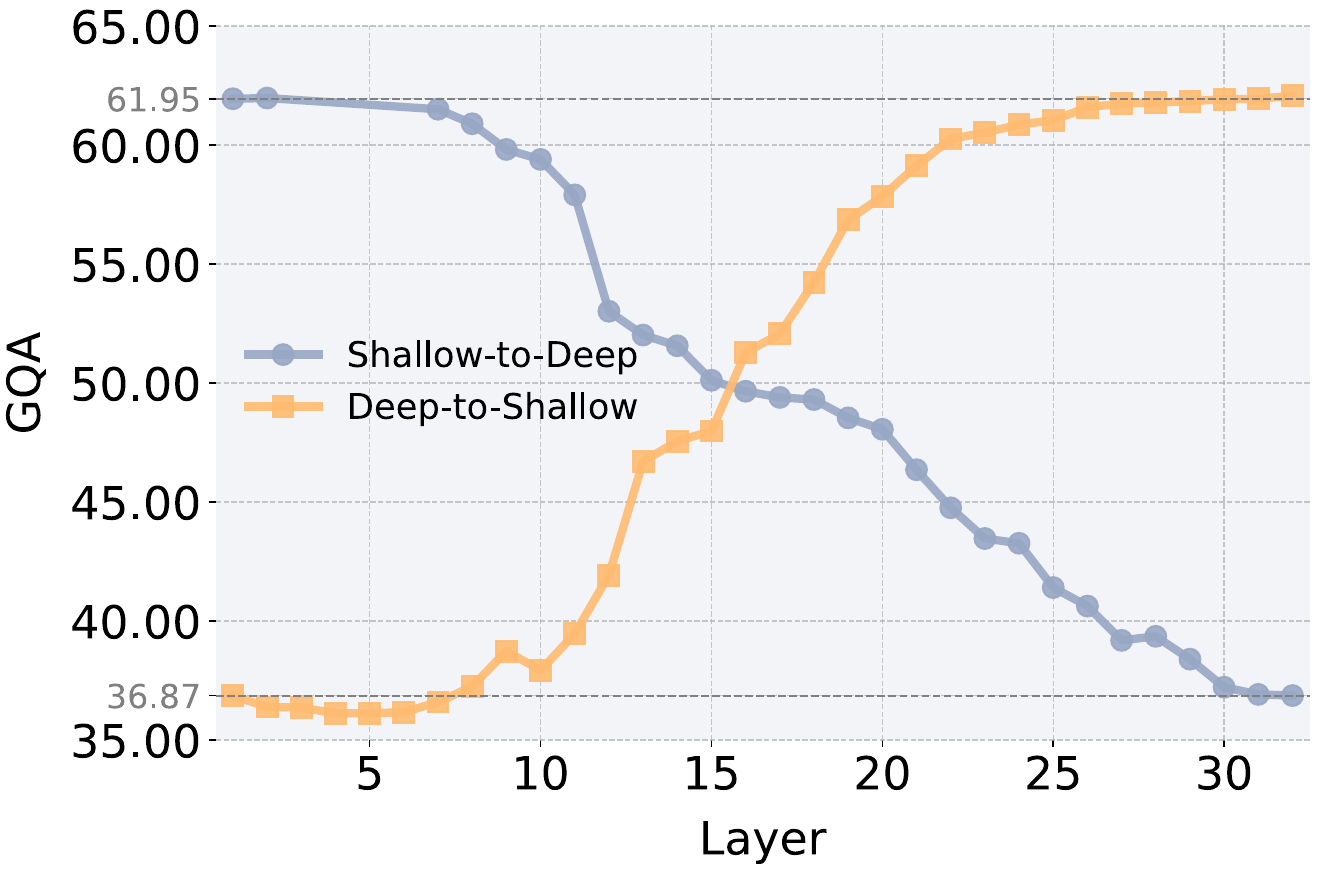}
    \vspace{-20pt}
    \caption{\small Masking ranges of layers, from shallow-to-deep and deep-to-shallow, exhibit a clear reduction in cross-modal fusion at both shallow and deep layers.}
    \label{fig:llava_line_chart}
\end{figure}

\section{Middle Layers: Sparse Grounding}
\label{sec:mid}

Beyond certain stage, we find that fully masking cross-attention begins to significantly deteriorate performance again from around 9th layer as shown in \autoref{fig:llava_line_chart}, suggesting a transition into middle layers. 

\subsection{Confirming Cross-Modal Fusion}
\label{subsec:mid_doing}
 
Given our prior analysis of shallow layers, this performance drop may also result from disruptions in the attention distribution rather than cross-modal interaction, so we perform two key analyses: \textit{attention merging and key visual token masking.}

\paragraph{Re-examine Attention Merging} 
We examine the impact of attention merging (\autoref{subsec:shallow_merging}) in middle layers 
. Compared to simple cross-attention masking, attention merging results in worse performance with GQA on LLaVA-v1.5 7B: \( 61.95 \to 51.73 \to 49.42 \), suggesting that the drop is not merely due to attention distribution disruption.

\paragraph{Key Visual Token Masking}
Next, we examine whether middle-layer attention is instruction relevant. We mask the top and bottom 10\% attended visual tokens for comparison in layers 9–15: 

\begin{itemize}[leftmargin=*, labelsep=0.5em, itemsep=0pt,parsep=0.8pt]
    \item \textbf{Top 10\% tokens}: GQA \( 61.95 \to 54.09 \)
    \item \textbf{Bottom 10\% tokens}: GQA \( 61.95 \to 61.93 \)
\end{itemize}

The significant performance drop when masking highly attended visual tokens, compared to the negligible impact of masking least-attended tokens, suggests that in the middle layers, cross-attention is focused on instruction-relevant regions, \emph{confirming meaningful cross-modal fusion} in these layers.

\subsection{Sparsity of Cross-Modal Fusion}
Given that middle layers are fusing visual features, we explore this fusion requires all visual tokens or only a sparse subset of them.

\paragraph{Selective Vision Masking} 
We apply cross-attention-based selection, retaining only the top 5\% most attended tokens unmasking, discarding remaining 95\%. The model still maintains a comparable performance (\( 72.6 \to 71.3 \)), confirming that middle layers start to focus on a sparse subset of vision tokens, rather than the entire image.

\paragraph{Visual Focus Tracking} While each middle layer may shift its focus to different visual regions when searching for the answer, we visualize the locations of critical visual tokens on the image and find that the model consistently focuses on instruction-relevant regions across layers (see \autoref{app:vis_mid_attn_based}).

These results imply that (1) \emph{cross-modal fusion in middle layers is sparse, only a few critical visual tokens are required}; and (2) \emph{critical visual tokens stay unchanged across layers, there is no need to re-identify critical tokens at each layer}. 

\subsection{Identifying Critical Visual Tokens}

Regarding this sparsity, we aim to develop a method that accurately identifies these critical visual tokens to reduce complexity. The most straightforward method is based on cross-attention weights. However, we find this approach is limited by (1) \emph{Visual Attention Sink Tokens}: The visual attention sink phenomenon is present across all layers, introducing irrelevant tokens in attention-based selection; (2) \emph{Difficulty Isolating Single Token Influence}: Attention weights are distributed across all tokens, which can introduce uncertainty when isolating the impact of individual tokens; and (3) \emph{Static Thresholds on Tokens Number}: Attention-based selection requires setting a fixed threshold, which reduces flexibility across different tasks.

Another intuitive approach to measure the influence of each vision token is to mask them individually and observe their effect on the final output. However, this requires propagating changes through all layers, making it computationally expensive. Instead, we propose a more efficient method that directly evaluates the impact of each vision token on the attention output of the last input token, which determines the first answer token.

\paragraph{Attention Computation Recap}
The attention weight matrix is calculated as:
\vspace{-5pt}\begin{equation}
    W = \mathrm{softmax}(\frac{QK^T}{\sqrt{d_k}}+M)
\end{equation}
where $Q, K$ are the query and key matrices, $W$ is the attention weight,and $M$ is a causal mask. The attention output is then computed by:
\vspace{-5pt}\begin{equation}
    O = \mathrm{Reshape}(\sum_{heads} W \cdot V)
\end{equation}
where $V$ is the value matrix, $O$ the attention output. 

\paragraph{Token Masking Procedure}
To evaluate the influence of token $j$ on token $i$ at layer $\ell$, we modify the attention weight matrix as follows:
\vspace{-5pt}\begin{equation}
    W_{i\rightarrow j}^{\prime} = 0 \label{f:maskAttnW}
\end{equation}
which masks the ability of token $i$ to attend to token $j$ across all attention heads. Using this masked attention weight, we recompute the attention output:
\vspace{-5pt}\begin{equation}
    O^{\prime}= \mathrm{Reshape}(\sum_{heads} W_{i\rightarrow j}^{\prime} \cdot V)
\end{equation}

\paragraph{Influence Measurement}
The influence of token $j$ on token $i$ is quantified by comparing the original attention output of token $i$ and the masked attention output of token $i$ using two complementary metrics: cosine similarity and L2 distance.

We measure the directional similarity between the original and masked outputs:
\vspace{-6pt}\begin{equation*}
    \text{Cosine Similarity}_{i\leftarrow j}=\frac{O_i\cdot \ O_{i\ \mathrm{masked}}^{\prime}}{\|O_i\|_2\ \| O_{i\ \mathrm{masked}}^{\prime}\|_2}.
    \label{f:cossimi}
\end{equation*}

where $\| \cdot \|_2$ is the L2-norm.
A lower similarity indicates a stronger influence of token $j$ on token $i$, as masking token $j$ significantly alters the output.

In addition to directional changes, we also measure the magnitude of change using the L2 distance:
\vspace{-6pt}\begin{equation} 
    \text{L2 Distance}_{i \leftarrow j} = | O_i - O_{i, \mathrm{masked}}^{\prime} |_2. 
    \label{f:l2norm} 
\end{equation}
A larger L2 distance reflects a greater impact of token $j$ on token $i$, as it quantifies the absolute difference in output magnitude after masking. 

By combining cosine similarity and L2 distance, we capture both directional and magnitude-based influences of vision tokens, offering a better way to identify \textbf{the most critical tokens} than using attention weights (See~\autoref{tab3:detailed_comparison} for detailed comparison).

\subsection{Strategy for Efficient MLLM Design}
\label{subsec:mid_pruning}
Given the sparsity of cross-modal fusion in middle layers, we propose an adaptive, training-free pruning strategy that retains only the most influential vision tokens: If masking a vision token reduces the cosine similarity below 0.995, we define this layer as a filtering layer, implying the visual input starts to contribute to the answer generation. Then, at this filtering layer, we discard vision tokens with a L2 distance below 0.2, as they have a negligible impact on the last input token. Using this method, we prune 576 vision tokens down to an average of 10.3 
after the filtering layer, maintaining competitive performance with only a 0.7\% drop in GQA. Moreover, our middle-layer pruning offers a new interpretability lens on vision token redundancy by lowering the minimum visual tokens retained.

\begin{table}[h!]
\centering \footnotesize
    \begin{tabular}{ccccc}
    \toprule
        \textbf{Strategy} & \textbf{POPE} & \textbf{GQA} & \textbf{\textbf{VQA}\textsuperscript{T}} & 
        \textbf{MMVet}\\
    \midrule
        Attn\footnotemark (last) & 85.9 & 60.3 & 57.1 & 25.4 \\
        Attn (text) & 85.9 & 58.0 & 55.6 & 23.8\\
        Attn (vis) & 85.9 & 55.2 & 52.0 & 20.9 \\
        \cellcolor{lightGray!30}Value-aware  & \cellcolor{lightGray!30}\textbf{86.1} & \cellcolor{lightGray!30}\textbf{61.3} & \cellcolor{lightGray!30}\textbf{57.8} & 
        \cellcolor{lightGray!30}\textbf{31.9} \\
        
    \bottomrule
    \end{tabular}
\vspace{-5pt}
\caption{Value-aware pruning in middle layers consistently outperformances attention-based methods, particularly in multi-token generation tasks like GQA, TextVQA and MMVet, indicating a stronger ability to retain instruction-relevant visual information.}
\label{tab3:detailed_comparison}
\end{table}
\footnotetext{Top 10 visual tokens most attended by the final text token, instruction tokens and visual tokens at layer 16.}
\begin{table*}[t!]
    \centering 
    \footnotesize
    \begin{tabular}{l@{\hspace{6pt}} cc|c@{\hspace{6pt}}c@{\hspace{6pt}}c@{\hspace{6pt}}c@{\hspace{6pt}}c@{\hspace{6pt}}c@{\hspace{6pt}}c|c}
        \toprule
            \textbf{Models} & \textbf{FLOPs(T)}
                & \textbf{Method} & \multicolumn{1}{c}{\textbf{GQA}} & \textbf{SQA}\textsuperscript{I} & \textbf{VQA}\textsuperscript{T} & \textbf{POPE} & \textbf{MME}\textsuperscript{P} & \textbf{MMB} & \textbf{MMStar} & \textbf{Avg. $\uparrow$} \\
        \midrule
            \multirow{2}{*}{\textbf{LLaVA-v1.5 7B}} 
                 & 3.82 & \cellcolor{lightGray!30}dense   & \cellcolor{lightGray!30}62.0 & \cellcolor{lightGray!30}66.8 & \cellcolor{lightGray!30}58.2 & \cellcolor{lightGray!30}85.9 & \cellcolor{lightGray!30}1507.6 & \cellcolor{lightGray!30}64.3 & \cellcolor{lightGray!30}33.7 & \cellcolor{lightGray!30}63.8 \\
                & 1.76 & ours     
                & 60.3 & 66.7 & 55.2 
                & 84.4 & 1428.3 & 62.0 
                & 33.3 & 61.9 \\
            \midrule
            \multirow{2}{*}{\textbf{LLaVA-v1.5 13B}}
                & 7.44 & \cellcolor{lightGray!30}dense   & \cellcolor{lightGray!30}63.3  & \cellcolor{lightGray!30}71.6 & \cellcolor{lightGray!30}61.3 & \cellcolor{lightGray!30}85.9 & \cellcolor{lightGray!30}1531.3 & \cellcolor{lightGray!30}67.7 & \cellcolor{lightGray!30}36.2 & \cellcolor{lightGray!30}66.1\\
            
                & 3.31 & ours     
                & 61.3 & \textbf{72.0} & 59.1 
                & 84.7 & 1485.3 & 66.9 
                & 36.0 & 64.9
                \\
            \midrule
            \multirow{2}{*}{\textbf{InternVL-v2.5 8B}}
                & 11.00 & \cellcolor{lightGray!30}dense   & \cellcolor{lightGray!30}63.6 & \cellcolor{lightGray!30}98.0 & \cellcolor{lightGray!30}79.1 & \cellcolor{lightGray!30}90.6 & \cellcolor{lightGray!30}1680.8 & \cellcolor{lightGray!30}84.6 & \cellcolor{lightGray!30}60.4 & \cellcolor{lightGray!30}80.1 \\
            
                & 5.34 & ours   
                & 58.8 & 97.8 & 77.7
                & 88.2 & 1643.5 & 79.6 
                & 59.9 & 77.0 \\
            \midrule
            \multirow{2}{*}{\textbf{QwenVL-v2 7B}}
                & 9.62 & \cellcolor{lightGray!30}dense   & \cellcolor{lightGray!30}62.4 & \cellcolor{lightGray!30}85.4 & \cellcolor{lightGray!30}76.9 & \cellcolor{lightGray!30}87.9 & \cellcolor{lightGray!30}1687.7 & \cellcolor{lightGray!30}79.4 & \cellcolor{lightGray!30}56.3 & \cellcolor{lightGray!30}64.6 \\
            
                & 4.69 & ours   
                & 62.2 & 84.1 & 74.4 
                & 87.8 & 1615.8 & 78.0 
                & 77.9 & 63.5 \\
            \midrule
            \multirow{2}{*}{\textbf{MobileVLM-v2 3B}}
                & 0.37 & \cellcolor{lightGray!30}dense & \cellcolor{lightGray!30}61.0 & \cellcolor{lightGray!30}70.0 & \cellcolor{lightGray!30}57.5 & \cellcolor{lightGray!30}84.7  & \cellcolor{lightGray!30}1440.5 & \cellcolor{lightGray!30}63.2 & \cellcolor{lightGray!30}35.1 & \cellcolor{lightGray!30}63.5 \\
            
                & 0.25 & ours
                & 57.6 & 69.4 & 53.5 
                & 81.7 & 1402.3 & 57.4
                & \textbf{36.7} & 63.3 \\
        \bottomrule
    \end{tabular}
    \vspace{-5pt}
    \caption{ \textbf{Performance of \emph{VisiPruner} across various MLLMs and benchmarks.} These benchmarks include visual question answering datasets GQA~\cite{benchmark:GQA}, MME~\cite{benchmark:mme}, MMBench~\cite{benchmark:mmbench}, and MMStar~\cite{benchmark:mmstar},  visual reasoning benchmark SQA~\cite{benchmark:sqa}, OCR benchmark TextVQA~\cite{benchmark:textvqa}, and the object hallucination benchmark POPE~\cite{benchmark:pope}.}
    \label{tab:ablation-results} 
\end{table*}

\begin{table*}[t!]
    \centering 
    \footnotesize
    \renewcommand{\arraystretch}{1.1}
    \begin{tabular}{l@{\hskip 4pt}c@{\hspace{6pt}} | ccccccc | c}
    \toprule
        \textbf{Method} & \textbf{Vis Attn Computation} & \textbf{MMB} & \textbf{\textbf{SQA}\textsuperscript{I}} & \textbf{GQA} & \textbf{MME}\textsuperscript{P} & \textbf{VQA}\textsuperscript{T} & \textbf{POPE} & \textbf{MMVet} & \textbf{Avg.} \\
    \midrule
        LLaVA-v1.5 7B & 100.0\% & 64.3 & 66.8 & 62.0 & 1507.6 & 58.2 & 85.9 & 31.2 & 63.4 \\
    \midrule
        PDrop\textsubscript{ retained=192} & $-86.4\%$ & 63.2 & 70.2 & 57.1 & 1419.8 & 56.1 & 82.3 & 30.5 & 61.5 \\
        SparseVLM\textsubscript{ retained=192} & $-86.4\%$ & 64.1 & 68.7 & 59.5 & 1441.1 & 56.1 & 85.3 & 33.1 & 62.7 \\
        FastV\textsubscript{k=3,r=0.75} & $-87.3\%$ & 63.5 & 68.7 & 57.5 &  1458.9 & 56.2 & 81.0 & 27.9 & 61.1 \\
    \midrule
        PDrop\textsubscript{ retained=64} & $-97.6\%$ & 33.3 & 69.2 & 41.9 & 982.3 & 45.9 & 55.9 & 30.7 & 46.6 \\
        SparseVLM\textsubscript{ retained=64} & $-97.6\%$ & 60.1 & 69.8 & 53.8 & 1351.4 & 53.4 & 77.5 & 24.9 & 58.2 \\
        FitPrune\textsubscript{ reduction=0.9} & $-98.0\%$ & 55.4 & 67.8 & 52.4 & 1210.2 & 52.1 & 60.5 & 24.2 & 53.3 \\
    
        \cellcolor{lightGray!30}Ours & \cellcolor{lightGray!30}$-98.3\%$ & \cellcolor{lightGray!30}62.0 & \cellcolor{lightGray!30}66.7 & \cellcolor{lightGray!30}60.3 & \cellcolor{lightGray!30}1428.3 & \cellcolor{lightGray!30}55.2 & \cellcolor{lightGray!30}84.4 & \cellcolor{lightGray!30}29.1
        & \cellcolor{lightGray!30}61.3 \\
    
    \bottomrule
    \end{tabular}
    \vspace{-5pt}
    \caption{ \textbf{Compare \emph{VisiPruner} with training-free token-wise compression baselines}, including: \textbf{FastV} \cite{chen2024imageworth12tokens}, which keeps tokens selected by the last-to-vision attention; \textbf{FitPrune} \cite{ye2024fitprunefasttrainingfree}, which prunes tokens according to attention-distribution saliency; \textbf{SparseVLM} \cite{zhang2025sparsevlmvisualtokensparsification}, which drops tokens based on cross-attention importance; and \textbf{PyramidDrop} \cite{xing2024pyramiddropacceleratinglargevisionlanguage}, which progressively reduces visual tokens.}
    \label{tab:ablation-results} 
\end{table*}

\section{Deep Layers: Linguistic Alignment}
\label{sec:deep}

As seen in Figure~\ref{fig:llava_line_chart}, we observe that beyond certain layers, masking all cross-attention connections once again has minimal impact on performance, which indicates a transition to deep layers.

\subsection{Discontinuous Vision Dependence}

To explore the role of vision tokens in different layers, we compare the performance impact of discarding visual tokens versus skipping visual processing only at specific layers. This allows us to better understand when vision tokens can be discarded.

\paragraph{Skipping $\neq$ Discarding}

When we discard all visual tokens from layer 20 and beyond, we observe a noticeable drop in performance on the GQA dataset, from \( 61.95 \) to \( 59.13 \). However, when we only skip the visual processing at layer 20 and allow visual information to continue through subsequent layers, the performance degradation is minimal, from \( 61.95 \) to \( 61.66 \). This suggests that while visual tokens remain relevant beyond layer 20, the processing in this layer itself is not essential. Therefore, we conclude that \emph{vision dependence may not be continuous}. Specifically, skipping one layer of visual processing does not necessarily imply that skipping all subsequent layers yields the same.

\paragraph{Discarding in Deep Layers}

Next, we investigate the impact of discarding visual tokens from layer 26. On GQA, we observe negligible performance change, from \( 61.95 \) to \( 61.91 \), indicating that the visual information processed in earlier layers is already sufficiently integrated. However, when we skip visual processing at layer 26 and allow subsequent layers to process the visual information, the performance drops more significantly, from \( 61.95 \) to \( 61.40 \). This suggests that by layer 26, visual tokens have already been integrated into the textual representation, and the visual information starts to introduce noise or redundancy in later layers.

Further supporting this, we observe minimal performance loss when masking cross-attention in deeper layers (\autoref{fig:llava_line_chart}), as well as when removing vision from the deep-layer KV Cache (\autoref{app:decoding}). These results reinforce the idea that \emph{after a certain layer, vision tokens can be safely discarded without significantly affecting performance}.
\subsection{Behavior: Linguistic Alignment} 

\begin{figure}[t]
    \flushleft 
    \begin{minipage}[t!]{0.44\columnwidth} 
        \includegraphics[width=\linewidth]{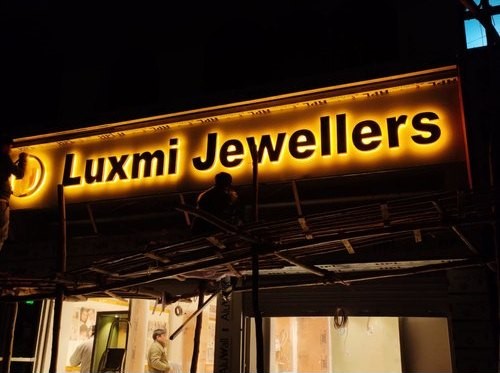}
    \end{minipage}%
    \hspace{0.5mm}
    \begin{minipage}[t!]{0.4\columnwidth} 
        \centering
        \tiny
        \renewcommand{\arraystretch}{0.94}
        \begin{tabular}{c@{\hspace{0.5mm}}l}
        \toprule
            \textbf{Layers} & \textbf{Top words in vocabulary space}\\
        \midrule
            32  & \textcolor{customGreen}{\textbf{The}}, In, All, """", There, \textbf{L}, \textbf{Lux}, I \\& A, It, \textbf{Lux} \\
            31  & \textcolor{customGreen}{\textbf{The}}, All, In, """", There, \textbf{L}, the, all \\& A, It, \textbf{Lux} \\
            30  & All, \textcolor{customGreen}{\textbf{The}}, all, \textbf{Lux}, the, In, \textbf{lux}, \textbf{L} \\& A, It, There \\
            25  & \textbf{Lux}, \textbf{lux}, all, scene, the, scene, Scene \\& \textcolor{customGreen}{\textbf{The}},  A, It, There \\
        \bottomrule
        \end{tabular}
    \end{minipage}
    \vspace{-5pt}
    \caption{\small Top vocabulary tokens from the semantic projection of the last input token at each layer.}
    \label{fig:luxmi_scene_text}
\end{figure}

Using the prompt "\textit{What are all the scene text in the image?}", we project the hidden state of the last input token to the semantic space (\autoref{eq:last_token_projection}). As shown in \autoref{fig:luxmi_scene_text}, by layer 25, the model generates the correct visual answer "\textit{Lux}", but struggles to structure it into a coherent response, "\textit{The scene text is '\textbf{Luxmi Jewellers}'}." While visual content is correctly identified, it is initially misplaced linguistically. As we move to deeper layers, the model gradually refines the output, prioritizing tokens "\textit{The}" to form a grammatically correct sentence.

These findings suggest that \textit{deep layers are responsible for aligning the generated content with natural language conventions}. 

\subsection{Strategy for Efficient MLLM Design}
Having known that deep layers no longer rely on vision tokens, we introduce a pruning strategy to detect the completion of vision-to-text fusion: After identifying and the only retaining critical vision tokens from middle layers (\autoref{subsec:mid_pruning}), we continuously track their influence. If these kept tokens show no measurable impact for two consecutive layers, we define the latter layer as the \textbf{vision exit layer} ($\ell_{exit}$). Beyond $\ell_{exit}$, those retained vision tokens are removed, further eliminating redundant computations. On LLaVA-v1.5 7B, this method identifies an average vision exit at layer 23.9, while still maintaining the performance on GQA \( 62.0 \to 61.3 \to 61.0 \), confirming that deep layers operate independently of vision.

\section{$\mathcal{V}isi\mathcal{P}runer$ and Future MLLMs}
\label{sec:experiments}

Based on key insights into the role of vision tokens and cross-modal interactions within LLaVA-v1.5 7B, this section aims to (1) validate the generalization ability of our conclusions across diverse MLLMs and (2) provide actionable recommendations for future model design.

\paragraph{Generalization Ability}
We apply our analytical methods and pruning strategies to multiple MLLMs with different architectures, including LLaVA-v1.5 13B, MobileVLM-V2-3B~\cite{chu2024mobilevlmv2fasterstronger}, Qwen2-VL 7B~\cite{wang2024qwen2vlenhancingvisionlanguagemodels} and InternVL2.5-8B~\cite{ chen2025internvl25}.
InternVL2.5 and Qwen2-VL are recently released MLLMs that dynamically generates image tokens, allowing us to verify the scalability of our conclusions in models with more flexible visual processing. MobileVLM 3B is a compact model with significantly fewer image tokens, enabling us to test the applicability in a MLLM with less parameters.

\paragraph{Complexity Analysis} By eliminating visual-relevant attention in shallow layers and deep layers while adaptively pruning to 10 vision tokens in middle filtering layers, we reduce cross-modal attention operations to minimal levels, achieving $98.3\%$ reduction in visual-related attention computation and a $53.9\%$ reduction in FLOPs compared to baseline. Building on our vision-independent layer identification, we maintain only the most interactive vision tokens on average in middle layers while completely excluding visual tokens from KV caching in shallow and deep layers. This strategic retention reduces the original visual KV cache memory and further lowers computational overheads in long-sequence decoding scenarios. Details about FLOPs calculation are in \autoref{app:flops}.

\paragraph{Method Comparison} Given that our method disables visual attention in shallow layers, we use the visual attention FLOPs reduction ratio as the evaluation criterion to ensure a fair comparison. Notably, \emph{our layer-wise compression strategy is compatible with token pruning approaches} and can further reduce computational overhead through shallow-layer visual attention merging and early vision exit.

\paragraph{Suggestions for Future MLLMs}

Based on our findings, we propose several guidelines to improve the efficiency and interpretability of future MLLMs: \textit{(a) Truncate shallow visual layers and eliminate cross/self-attention} Since shallow layers contribute little to cross-modal fusion, computational overhead can be reduced by processing visual tokens only up to the middle layers. The model can be trained to recognize the start of middle layers, or adapted to a fixed starting point. \textit{(b) Train models to attend sparsely} By training for sparse attention in middle layers, the model directly identifies critical tokens, bypassing the need for post-hoc attention scores or influence measurements. \textit{(c) Enable early exiting in deep visual layers once modality fusion is established.} Given the established linguistic alignment behavior in deep layers, we recommend incorporating vision exit mechanisms into MLLM training pipelines to automatically skip out when fusion is finished.
\section{Conclusions}
\label{sec:con}

We propose a three‐stage MLLM framework—where shallow layers handle intra‐modal task interpretation, middle layers integrate task‐relevant visual tokens into textual embeddings, and deep layers focus on linguistic alignment. Building on these insights, we introduce stage-specific optimizations that boost computational efficiency, and validated our framework across multiple MLLM architectures, confirming its general applicability. Finally, we distill our findings into practical guidelines for future MLLM design.

\section*{Limitations}

While our study provides a principled and general framework for understanding the mechanisms of vision-language models, there are several limitations. First, training the projector to align vision tokens with semantic representations and inserting them until later layers could further strengthen our findings regarding intra-modal processing in shallow layers. Second, due to hardware constraints, our analysis was limited to models with up to 13 billion parameters. Future work could replicate our approach using larger models, potentially uncovering additional insights through our three-stage analytical framework.

\section*{Acknowledgement}

We thank EIT and IDT High Performance Computing Center for providing computational resources for this project.
This work was supported by the 2035 Key Research and Development Program of Ningbo City under Grant No.2024Z123 and No. 2025Z034.

\newpage
\bibliography{custom}
\newpage
\appendix

\section{Related Work}
\subsection{Cross-modal Information Flow in MLLMs}
Research on cross-modal information flow in MLLMs has shown that visual information is gradually integrated into the generation of subsequent textual tokens ~\cite{neo2024interpretingvisualinformationprocessing, wu2024acceleratingmultimodallargelanguage,zhang2024crossmodalinformationflowmultimodal,tong2025llmeffectivestreamingprocessor}. However, there is still disagreement about how and when this fusion occurs within the model.
\citet{neo2024interpretingvisualinformationprocessing} suggest that key visual information is primarily extracted in the middle to late layers of the model. In contrast, based on attention weight analysis, \citet{wu2024acceleratingmultimodallargelanguage} and \citet{zhang2025llavaminiefficientimagevideo} argue that visual information is fused into textual tokens in the shallow layers, highlighting the role of vision tokens early in the process. Similarly, \citet{zhang2024crossmodalinformationflowmultimodal} report that the model is constantly fusing visual information, starts with perceiving the entire image and then extracting key visual details.

\subsection{In-VLM Vision Compression}

Identifying and retaining important tokens that are crucial for generation is a key aspect of effective training-free token pruning ~\cite{xiao2024efficientstreaminglanguagemodels,zhang2023h2oheavyhitteroracleefficient,liu2023scissorhandsexploitingpersistenceimportance}. To make vision compression more adaptive to user instructions, in-VLM compression has become a key area of research. \citet{chen2024imageworth12tokens} observe the significant of redundancy of vision tokens via the sparsity of attention for vision tokens within VLMs, and propose a pruning method named FastV to pick the most important vision tokens based on attention each vision token received from the last token. Building on FastV, PyramidDrop drops vision tokens in multiple stages \cite{xing2024pyramiddropacceleratinglargevisionlanguage}. SparseVLM selects visual-relevant text tokens to evaluate the importance of vision tokens based on the self-attention matrix, then prunes the vision tokens using a rank-based strategy and token recycling to maximize sparsity while retaining essential information \cite{zhang2024sparsevlmvisualtokensparsification}.


\section{Mask Highly Attended Visual Tokens in Shallow Layers Using Different Selection Criteria}
\label{app:shallow-attended_masking}

To further validate that highly attended visual tokens has no effects, we conducted experiments on additional selection criteria:

\begin{table}[h!]
\centering \footnotesize
    \begin{tabular}{ccccc}
    \toprule
        \textbf{Criterion}   & \textbf{GQA} & \textbf{MME\textsuperscript{P}} & \textbf{VQA}\textsuperscript{T} & \textbf{POPE}\\
    \cmidrule(lr){1-1}\cmidrule(lr){2-5}
        \cellcolor{lightGray!30} vanilla & \cellcolor{lightGray!30}62.0   & \cellcolor{lightGray!30}1507.6  & \cellcolor{lightGray!30}58.2 & \cellcolor{lightGray!30}85.9 \\

        attn (last $\to$ vis) & 62.0 & 1506.6 & 57.9 & 85.7 \\
        attn (text $\to$ vis) & 62.0 & 1503.6 & 58.1 & 85.7 \\
        pos (near text)   & 62.0 & 1501.1 & 58.1 & 85.7 \\
    
    \hline
    \end{tabular}
\caption{Performance after masking top 60 attended visual tokens in the first two layers using different selection criteria.}
\label{tab0:masking_top_tokens_in_shallow_layers}
\end{table}


\section{Comparison of Cross-Attention Masking Across Different Stages}
\label{app:shallow-middle-deep}

We also compare the performance on different different benchmarks with cross attention masked in different stages as shown in \autoref{apptab:crossattnmasked}. The shallow and deep layers exhibit significantly cross-modal information fusion compared with middle layers.

\begin{table}[h!]
\centering \footnotesize
    \renewcommand{\arraystretch}{1.1}
    \begin{tabular}{ll | ccc}
    \toprule
        \textbf{Model} & \textbf{Layers} & \textbf{GQA} & \textbf{MME$^{P}$} & \textbf{VQA$^{T}$} \\
    \midrule
    \multirow{4}{*}{\textbf{LLaVA-v1.5 7B}} &
        \cellcolor{lightGray!30}Dense & \cellcolor{lightGray!30}62.0  & \cellcolor{lightGray!30} 1507.6 &  \cellcolor{lightGray!30}58.2\\
        & 1--7   & 61.5 & 1411.2 & 56.8 \\
        & 9--15  & 51.7 & 722.6  & 51.1 \\
        & 27--32 & 61.8 & 1488.5 & 58.1 \\
    \bottomrule
    \end{tabular}
\caption{\label{apptab:crossattnmasked}Performance on Various Benchmarks with Cross-Attention Masked in Specific Layers.}
\end{table}


\section{Visualization of visual attention sink phenomenon} 
\label{app:vis_sink_token}

\begin{figure*}[t!]
    \centering
    \includegraphics[width=1\linewidth]{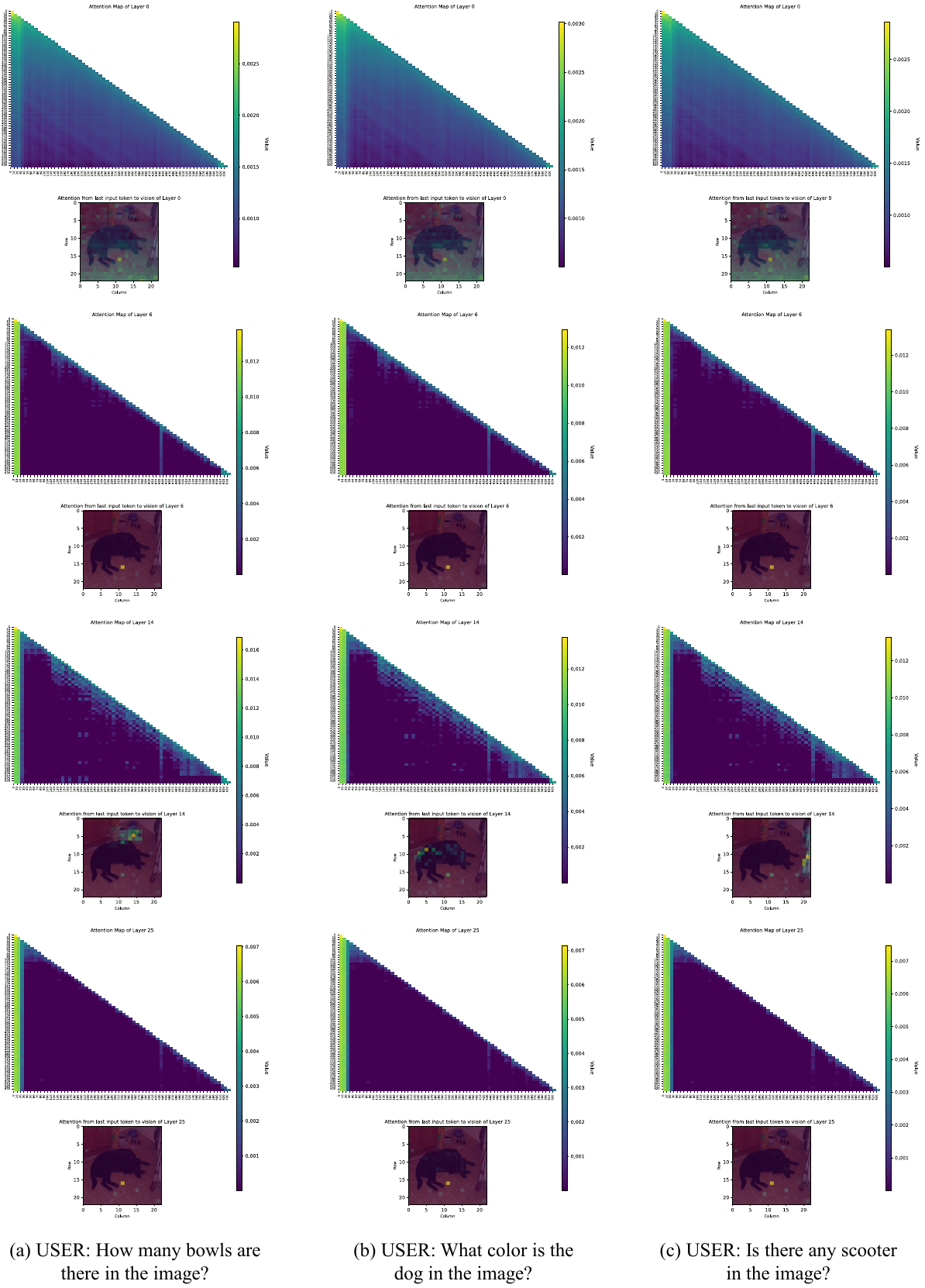}
    \caption{Visualization of attention map and distribution on image with different instruction across shallow, middle and deep layers using LLaVA-v1.5 7B}
    \label{fig:visualization_attention_distribution}
\end{figure*}

In \autoref{fig:visualization_attention_distribution}, we visualize the attention distribution on the input image across shallow, middle and deep layers to highlight the visual attention sink phenomenon. Ideally, attention distribution should adapt dynamically based on the input, directing focus to different areas for different tasks. However, our visualizations reveal an intriguing pattern: tokens with high attention scores—highlighted in the image—tend to appear consistently in the same regions across various instructions \textbf{in both shallow and deep layers}. This finding suggests that certain vision tokens act as attention sinks, drawing focus but failing to provide meaningful contributions to the model's reasoning. As a result, these tokens may not be essential for generating accurate responses.

Moreover, in the middle layers, we observe that the model starts to concentrate its attention on the more instruction-relevant areas. This reinforces our conclusion that MLLMs undergo a three-stage information processing approach, where shallow layers focus on task recognition, middle layers selectively fuse instruction-relevant visual information, and deep layers refine and align the response with the instruction.

Another interesting finding is that the first layer exhibits clear attention window, the lower half of vision tokens receive more attention from the last input token.

\section{Detailed Analysis on Visual Attention Sink Tokens}
\label{app:analysis_vis_sinks}

\subsection{Lower L1 Norm of Value Vectors for Sink Tokens}
As shown in the lower subplot of \autoref{fig:value_and_attn_weight}, visual sink tokens with high attention weights exhibit significantly lower magnitudes in their value vectors. This suggests that visual sink tokens function similarly to textual sink tokens, acting as bias terms in the softmax computation.

\subsection{Attention Redistribution After Removing Visual Sink Tokens}

After identifying the visual sink tokens in an example, we remove these tokens before the first layer. We observe that the attention weight previously allocated to the visual sink tokens is redistributed to the textual sink tokens in the system prompt.

\begin{figure}[H]
    \centering
    \includegraphics[width=1\linewidth]{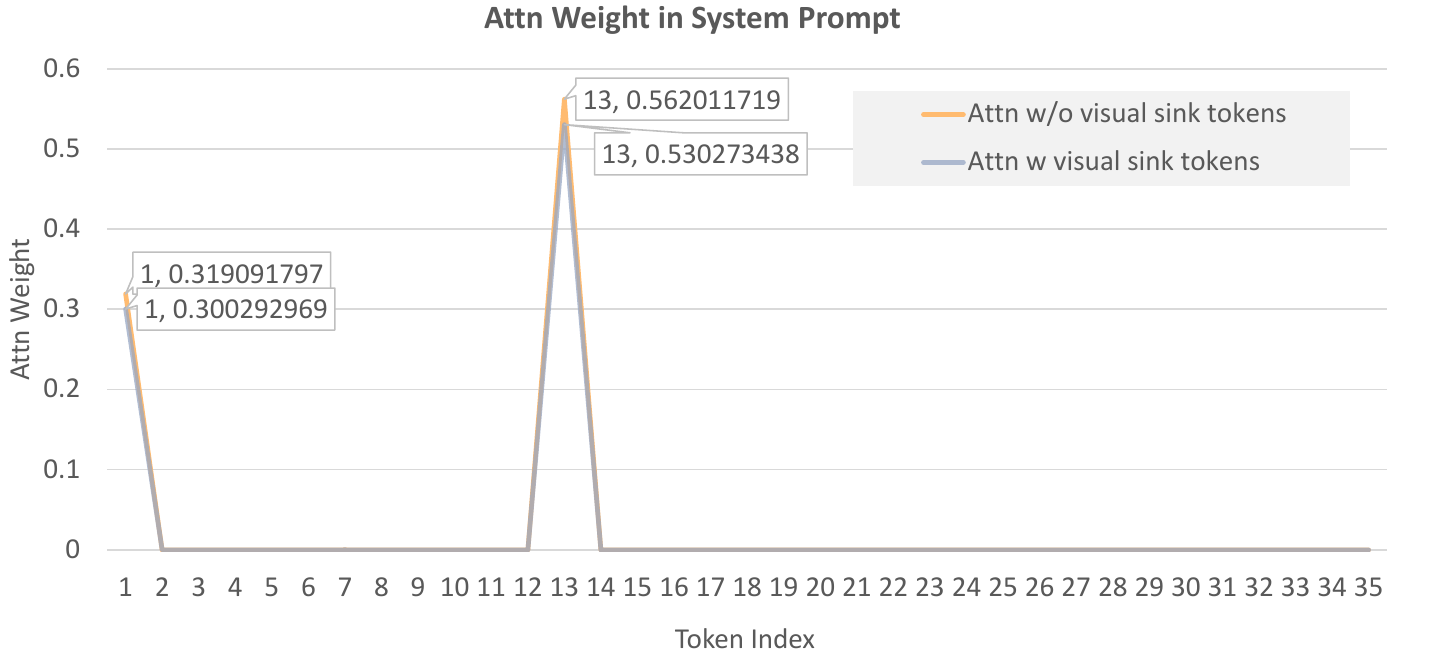}
    \caption{Textual sink tokens in the system prompt absorb the attention weight when visual sink tokens are removed in the third layer.}
    \label{fig:attn_weight_absorbed}
\end{figure}

Sum of attention weight from visual sink tokens: 0.053352.
Difference in attention weight of textual sink tokens with and without visual sink tokens: 0.050537109.

\section{L1 Norm of Value Vectors}
\label{app:l1_norm_value}

\begin{figure}[H]
    \centering
    \includegraphics[width=1\linewidth]{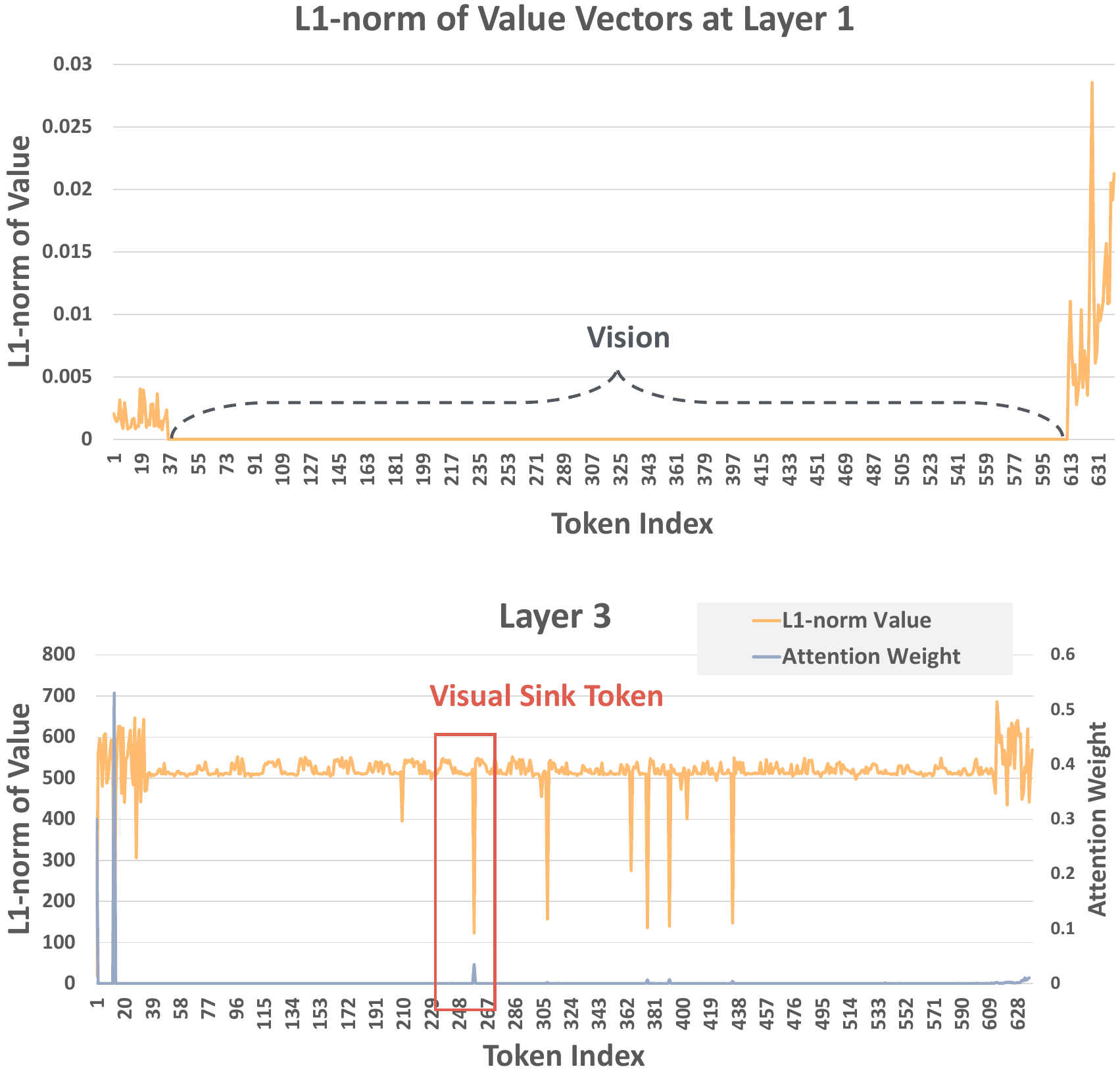}
    \caption{Visualization of attention map and distribution on image with different instruction across shallow, middle and deep layers using LLaVA-v1.5 7B}
    \label{fig:value_and_attn_weight}
\end{figure}

As illustrated in \autoref{fig:value_and_attn_weight}, the value vectors for textual and visual tokens show distinct patterns in the first layer. This likely indicates that the model differentiates between modalities at this stage, highlighting the necessity of modality-specific sinks.

\section{Random Selection of Visual Attention Merging Token}
\label{app:random_merging}

To ensure the visual token selection for merging is not index-dependent, we randomly choose a visual token and merge all visual cross-attention into it.

\begin{table}[h!]
\centering \footnotesize
    \begin{tabular}{cc}
    \toprule
        \textbf{Visual Token Index} & \textbf{GQA} \\
    \midrule
        \cellcolor{lightGray!30}vanilla & \cellcolor{lightGray!30}61.95 \\  
        - & 57.41 \\  
        1 & 61.98 \\  
        576 & 61.55 \\  
        128 & 61.83 \\  
        288 & 61.76 \\  
    \bottomrule
    \end{tabular}
\caption{Performance of random visual token merging on GQA.}
\label{app_tab:random_merging_token}
\end{table}

\section{Complete Results on Semantic Projection of the Last Input Token}
\label{app:projection}

In this section, we present a more detailed analysis of the semantic projection of the last input token for different user instructions.

\subsection{USER: How Many Cars Are in the Image?}

As shown in \autoref{app_tab:task_projection_count}, when given the user instruction "\textit{How many cars are there in the image?}", the model accurately identifies it as a number-related task.

\begin{table*}[t!]
\centering \footnotesize
    \renewcommand{\arraystretch}{1.1}
    \begin{tabular}{{ll|ccc|cc|c|c}}
    \toprule
        \multirow{2}{*}{\textbf{Model}} & \multirow{2}{*}{\textbf{Layers}} & \multicolumn{3}{c}{\textbf{Vision}} & \multicolumn{2}{c}{\textbf{Text}} & \multicolumn{1}{c|}{\textbf{Math}} & \multirow{2}{*}{\textbf{Overall}} \\
        & & Recognition  & OCR  & Spatial awareness & Knowledge & Generation & Math & \\
    \midrule
        \multirow{5}{*}{\textbf{LLaVA-v1.5 7B}} & 
        \cellcolor{lightGray!30}Dense & \cellcolor{lightGray!30}36.1 & \cellcolor{lightGray!30}23.9 & \cellcolor{lightGray!30}26.3 & \cellcolor{lightGray!30}17.1 & \cellcolor{lightGray!30}22.4 & \cellcolor{lightGray!30}11.5 & \cellcolor{lightGray!30}31.2 \\
        &0--7   & 39.5 & 25.2 & 28.8 & 21.4 & 26.9 & 15.4 & \textbf{33.8} \\
        &8--14  & 34.0 & 21.4 & 26.5 & 16.1 & 19.0 & 7.7  & 29.2 \\
        &25--31 & 35.9 & 22.2 & 23.2 & 18.6 & 22.4 & 11.2 & 31.1 \\
        &0--31  & 33.1 & 13.5 & 23.5 & 14.2 & 16.6 & 7.7  & 26.1 \\
    \bottomrule
    \end{tabular}
\caption{\textbf{Performance Breakdown of LLaVA-v1.5 7B on MM-Vet with Vision Removal from Specific Layers in the KV Cache.} "\textit{Layers}" column indicates the layers from which visual information was removed.}
\label{app_tab:mmvet}
\end{table*}

\begin{table}[h!]
\centering \footnotesize
    \begin{tabular}{cl}
    \toprule
        \textbf{Layers} & \textbf{ Top words in vocabulary space}\\
    \midrule
        19  & four, three, five, several, six, many, seven \\& two, Several, dozen \\
        18  & four, three, several, two, five, dozen, lots, many \\& \textcolor{customGreen}{\textbf{number}}, multiple \\
        17  & four, three, several, two, dozen, five, \textcolor{customGreen}{\textbf{number}} \\& mehrere, lots, multiple \\
        16  & four, three, \textcolor{customGreen}{\textbf{number}} \\& two, five, dozen, several\\& many, mehrere, lots \\
        15  & four, \textcolor{customGreen}{\textbf{number}}, three, Ges, dozen, several, lots\\& five, \textcolor{customGreen}{\textbf{count}}, multiple \\
        14  & four, \textcolor{customGreen}{\textbf{number}}, three, Ges, two, érique, \textcolor{customGreen}{\textbf{count}} \\& lots, There, ieri \\
        13  & \textcolor{customGreen}{\textbf{number}}, three, \textcolor{customGreen}{\textbf{count}}, \textcolor{customGreen}{\textbf{number}}, four, érique \\& none, ocker, multip, estaven \\
        12  & \textcolor{customGreen}{\textbf{number}}, arden, rita, \textcolor{customGreen}{\textbf{Number}}, multip, three\\& \textbf{NUM}, licz, \textcolor{customGreen}{\textbf{number}}, \textbf{NUM} \\
        11  & \textcolor{customGreen}{\textbf{number}}, arden, rita, \textcolor{customGreen}{\textbf{Number}}, none, licz \\& \textcolor{customGreen}{\textbf{number}}, Sa, three, Ges \\
        10  & \textcolor{customGreen}{\textbf{number}}, arden, rita, ubre, nim, konn, eben \\& multip, \begin{CJK}{UTF8}{gbsn}兴\end{CJK}, two \\
        9   & \textcolor{customGreen}{\textbf{number}}, rita, multip, nim, arden, platz, iken \\& zero, un, VS \\
    \bottomrule
    \end{tabular}
\caption{Top tokens from the projection of the last input token at each layer.}
\label{app_tab:task_projection_count}
\end{table}

\subsection{USER: What Kind of Apple Is This?}

As shown in \autoref{app_tab:task_projection_type}, when given the user instruction "\textit{What kind of apple is this?}", the model correctly identifies it as a type-related task.

\begin{table}[h!]
\centering \footnotesize
    \begin{tabular}{cl}
    \toprule
        \textbf{Layers} & \textbf{ Top words in vocabulary space}\\
    \midrule
        9 & sterd, publique, \textcolor{customGreen}{\textbf{typen}}, Hinweis, penas, ohl, bpe\\& Hero, Sob, ermeister \\
        8 & sterd, \textcolor{customGreen}{\textbf{typen}}, publique, paździer, \begin{CJK}{UTF8}{gbsn}庄\end{CJK}, schrift \\& \begin{CJK}{UTF8}{gbsn}泉\end{CJK}, intrag, penas, Hinweis \\
        7 & sterd, penas, quelle, \textcolor{customGreen}{\textbf{typen}}, \begin{CJK}{UTF8}{gbsn}泉\end{CJK}, teil, wohl \\& paździer, \begin{CJK}{UTF8}{gbsn}庄\end{CJK}, intrag \\
        6 & sterd, paździer, strij, sierp, kwiet, penas, ści \\& Wikispecies,   wohl, konn \\
    \bottomrule
    \end{tabular}
\caption{Top tokens from the projection of the last input token at each layer.}
\label{app_tab:task_projection_type}
\end{table}


\section{Task Recognition: Projection of Value-Output Matrix on Semantic Space}
\label{app:proj_vo}

The value-output matrix plays a key role in in-context learning by summarizing task-related information. Building on the approach from~\cite{dar2023analyzingtransformersembeddingspace}, we project this matrix into the semantic space as follows: 
\begin{equation}
    D=W_u(V_{last}\cdot O)
\end{equation}
where $V$ is the value vector, $O$ is the output matrix, and $W_u$ is the word unembedding matrix.

\subsection{USER: Where is the place of origin?}
Given the instruction "\textit{Where is the place of origin?}", the model recognizes this as a location-related task \autoref{app_tab:vo_projection_location}.

\begin{table}[H]
\centering \footnotesize
    \begin{tabular}{ccl}
    \toprule
        \textbf{Layer} & \textbf{Head} & \textbf{Top words in vocabulary space}\\
    \midrule
        14 & 31 & names,Names,NAME,ját,Names \\
        13 & 31 & \textcolor{customGreen}{location,locations,map,Location,Map} \\
        12 & 31 & thy,thee,thou,Gemeins,Tu \\
    \bottomrule
    \end{tabular}
\caption{Top 5 tokens from the semantic projection of the value-output matrix of the last input token at different layers.}
\label{app_tab:vo_projection_location}
\end{table}

\subsection{USER: How many apples are there in the image?}
Given the instruction "\textit{How many apples are there in the image?}", the model recognizes this as a counting-related task \autoref{app_tab:vo_projection_counting}.

\begin{table}[H]
\centering \footnotesize
    \begin{tabular}{ccl}
    \toprule
        \textbf{Layer} & \textbf{Head} & \textbf{Top words in vocabulary space}\\
    \midrule
        13 & 31 & two,another,deux,atori,three \\
        12 & 31 & \textcolor{customGreen}{counting,counts,numbers,count,count} \\
        11 & 31 & \begin{CJK}{UTF8}{gbsn}你\end{CJK},your,you,vous,yourself \\
    \bottomrule
    \end{tabular}
\caption{Top 5 tokens from the semantic projection of the value-output matrix of the last input token at different layers.}
\label{app_tab:vo_projection_counting}
\end{table}

\subsection{USER: What is the make of the car on the left?}
Given the instruction "\textit{What is the make of the car on the left?}", the model recognizes this as a brand-related task \autoref{app_tab:task_projection_brand}.

\begin{table}[H]
\centering \footnotesize
    \begin{tabular}{ccl}
    \toprule
        \textbf{Layer} & \textbf{Head} & \textbf{Top words in vocabulary space}\\
    \midrule
        14 & 31 & different,Wat,isse,iesen,newer \\
        13 & 31 & \textcolor{customGreen}{brand,companies,company,Brand,brand} \\
        12 & 31 & loro,ihnen,your,their,nx \\
    \bottomrule
    \end{tabular}
\caption{Top 5 tokens from the semantic projection of the value-output matrix of the last input token at different layers.}
\label{app_tab:task_projection_brand}
\end{table}


\section{Analysis of Vision Removal Impact on MM-Vet Performance in KV Cache} \label{app:mmvet_breakdown}

To further probe the role of shallow layers, we conducted a vision removal experiment using MM-Vet, a benchmark requiring extended responses where key visual information must be preserved in the KV Cache. Specifically, we examined whether the model relies on vision information from shallow layers during the decoding process. A detailed breakdown of MM-Vet with vision removal on specific layers to determine whether performance degradation or improvement is attributed to vision or text generation. After pruning visual information from the first eight layers, the model performed better than the original configuration, further consolidating that the model does not utilize visual information from shallow layers (see \autoref{app_tab:mmvet}). Additionally, removing vision tokens in deep layers also have a minimal influence on the performance, indicating that the model focuses on processing textual information to align with instruction.


\section{Visualization of Instruction-Relevant Focus Across Middle Layers}
\label{app:vis_mid_attn_based}

\begin{figure}[h!]
    \centering
    \includegraphics[width=0.8\linewidth]{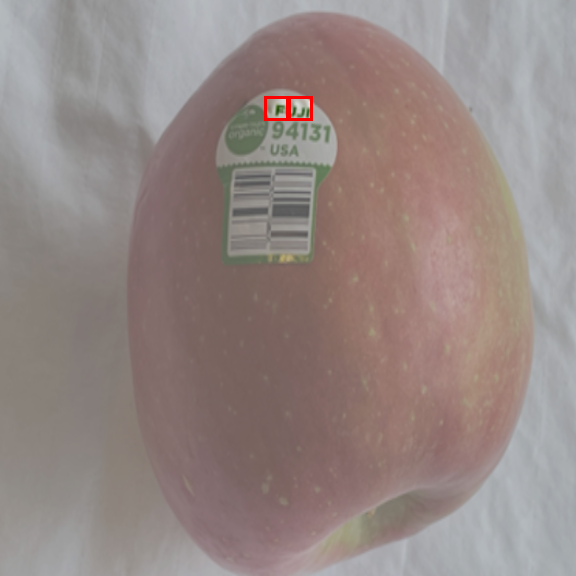}
    \caption{\textbf{The Most Instruction-Relevant Region Highlighted in Red Boxes.}}
    \label{fig:fuji_apple}
\end{figure}

Given the user instruction "\textit{What kind of apple is this?}" and the image in \autoref{fig:fuji_apple}, we observe that the last token in the middle layers consistently focuses on the most instruction-relevant region (see \autoref{app_tab:constant_focus}).

\begin{table}[h]
\centering \footnotesize
    \begin{tabular}{cl}
    \toprule
        \textbf{Layers} & \textbf{ Top 10 Visual Tokens Indices}\\
    \midrule
        22  & \textcolor{customGreen}{\textbf{107}}, \textcolor{customGreen}{\textbf{108}}, 129, 130,  60, \textcolor{customRed}{\textbf{222}}, 155, 255, 512, 162 \\
        21  & \textcolor{customGreen}{\textbf{107}}, \textcolor{customGreen}{\textbf{108}}, 129, 130,  60, \textcolor{customRed}{\textbf{222}}, 155, 255, 512, 162 \\
        20  & \textcolor{customGreen}{\textbf{107}}, \textcolor{customGreen}{\textbf{108}},  60, 162, 161, \textcolor{customRed}{\textbf{222}}, 163,  61, 399, 255 \\
        19  & \textcolor{customGreen}{\textbf{108}}, \textcolor{customGreen}{\textbf{107}},  60, \textcolor{customRed}{\textbf{222}}, 255, 387, 399,  61, 207, 299 \\
        18  & \textcolor{customGreen}{\textbf{108}}, \textcolor{customRed}{\textbf{222}}, \textcolor{customGreen}{\textbf{107}}, 207,  60, 502, 155,  88, \textcolor{customRed}{\textbf{355}}, 399 \\
        17  & \textcolor{customGreen}{\textbf{107}}, \textcolor{customRed}{\textbf{222}}, \textcolor{customGreen}{\textbf{108}}, 155,  60, 512, 130, 156, 255, 129 \\
        16  & \textcolor{customGreen}{\textbf{107}}, \textcolor{customGreen}{\textbf{108}}, \textcolor{customRed}{\textbf{222}}, 155,  60, 156, 131, \textcolor{customRed}{\textbf{355}}, 109, 340 \\
        15  & \textcolor{customGreen}{\textbf{107}}, \textcolor{customGreen}{\textbf{108}}, \textcolor{customRed}{\textbf{222}},  60,  61, 255,  88, 163, 399, 155\\
        14  & \textcolor{customRed}{\textbf{222}}, \textcolor{customGreen}{\textbf{107}}, \textcolor{customRed}{\textbf{355}}, \textcolor{customGreen}{\textbf{108}}, 340, 159, 574, 255, 398, 131 \\
        13  & \textcolor{customRed}{\textbf{222}}, \textcolor{customGreen}{\textbf{107}}, \textcolor{customRed}{\textbf{355}}, \textcolor{customGreen}{\textbf{108}}, 340, 398, 574, 255,  60, 155 \\
        12  & \textcolor{customRed}{\textbf{222}}, \textcolor{customRed}{\textbf{355}}, 340, 398, 270, 155, 574, \textcolor{customGreen}{\textbf{107}}, 272, 207 \\
        11  & \textcolor{customRed}{\textbf{222}}, \textcolor{customRed}{\textbf{355}}, 340, 574, 575, 398, \textcolor{customGreen}{\textbf{108}}, \textcolor{customGreen}{\textbf{107}}, 155, 156 \\
        10  & \textcolor{customRed}{\textbf{222}}, 575, \textcolor{customRed}{\textbf{355}}, 574, 340, 398, 207, 571, 272, \textcolor{customGreen}{\textbf{\textcolor{customGreen}{\textbf{108}}}} \\
    \bottomrule
    \end{tabular}
\caption{Top 10 most attended vision tokens from the last input token at each layer. \textcolor{customGreen}{Green} indicates the most critical visual tokens, while \textcolor{customRed}{red} marks the visual attention sink tokens.}
\label{app_tab:constant_focus}
\end{table}

\section{Layer-wise Cross-Attention Masking on MobileVLM 3B}
\label{app:cross_attn_masking_mobilevlm}

Compared to LLaVA-v1.5 7B, MobileVLM v2 3B has a broader range of shallow layers and fewer deep layers. This suggests that smaller models may require more computations on task recognition.

\begin{figure}[h]
    \centering
    \includegraphics[width=0.8\linewidth]{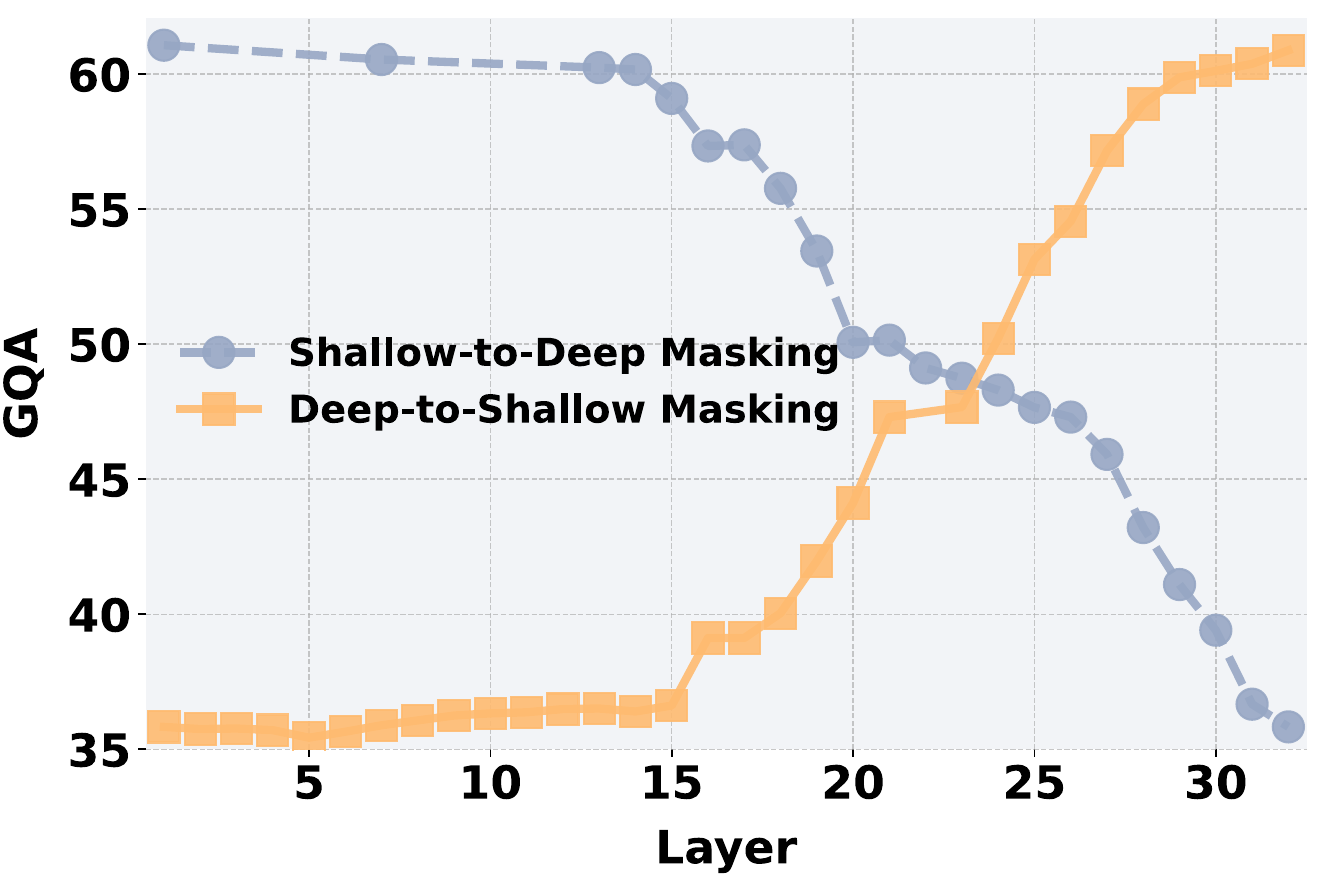}
    \caption{Impact of masking layer ranges from shallow-to-deep and deep-to-shallow, showing a clear reduction in cross-modal fusion in both shallow and deep layers.}
    \label{fig:mobileVLM_line_chart}
\end{figure}

\section{FLOPs Analysis on LLaVA-1.5 7B}
\label{app:flops}

Our proposed method greatly reduces vision-related self-attention, cross-attention and FFN, leading to an overall FLOPs reduction of $>60\%$. Here is a detailed analysis:

The total computation in MLLMs primarily consists of two components: attention computation and feed-forward network (FFN) computation. Among these, attention computation scales quadratically with sequence length, making it the primary computational bottleneck—especially in models like Qwen2-VL, which can generate up to 12,000 visual tokens. For instance, in LLaVA-1.5 7B, the FLOPs for attention computation can be expressed as $2n^2d$. The reduction ratio for visual attention computation is given by: 
\begin{equation*}
    R=1-\frac{L^{\prime}2*2(n_v^{\prime})^2d+L^{\prime}(n_v^{\prime}n_t)d}{32*(2(n_v^2d+n_vn_t))}
\end{equation*}
where the $L^{\prime}$ the number of cross-modal interaction layers, $n_v^{\prime}$ represents the number of retained visual tokens. If the input sequence consists of 650 tokens (576 visual tokens and 74 text tokens), our approach eliminates attention computation in shallow and deep layers, retaining only a few critical tokens for cross-modal fusion. This results in a 99\% reduction at maximum in attention computation. 

\paragraph{FLOPs Calculation.} In LLaMA 2 7B \cite{touvron2023llama2openfoundation}, the primary flops include FFN and self-attention. The flops for FFN is $3ndm$, where $n$ is the number of input tokens, $d$ is the hidden state size, and $m$ is the intermediate size of the FFN. Hence, the FLOPs overall calculation for visual tokens follows:
\begin{align*}
    &\sum_{i=0}^{L_{\text{middle}}} 
        \left(4n_v^{\prime}d^2 + 2n_v^{\prime2}d + 3n_v^{\prime}dm \right) \\
    &\quad + \sum_{i=0}^{L_{\text{shallow}}} 
        \left(4n_v d^2 + 3n_v dm \right)
\end{align*}
This optimization leads to an overall visual FLOPs reduction of 62.8\% under the given setting (576 visual tokens and 74 text tokens), significantly enhancing efficiency while maintaining performance. Given that the efficiency gain scales with longer textual or visual inputs, our pruning framework offers much greater benefits for longer text instructions or when multiple images are provided.

Additionally, following our actionable guidelines for optimizing MLLMs, the visual computation overhead within shallow layers in FFN should be able to be further reduced through training.

\section{Failure Case Analysis}
\label{app:sample}

In this section, we present an analysis on failure cases in GQA, where our pruned model produced 1,125 mismatched answers compared to the vanilla LLaVA-v1.5 7B over 12,000 samples.

\begin{itemize}
    \item 234 answers were correct in our model but incorrect in the vanilla model.
    \item 325 answers were incorrect in our model but correct in the vanilla model.
\end{itemize}

Upon closer inspection, we found that misclassifications were often related to \textbf{variations in word choice} rather than fundamental misunderstandings. Below are some examples:

\subsection{"Which kind of vehicle is in front of the flag?\textbackslash nAnswer the question using a single word or phrase."}
\begin{itemize}[leftmargin=*, labelsep=0.5em, itemsep=0pt]
    \item \textbf{Ground Truth Answer}: "van"
    \item \textbf{Vanilla Model}: "truck"
    \item \textbf{Ours}: "van"
\end{itemize}

\subsection{"What is sitting in front of the table that looks yellow and black?\textbackslash nAnswer the question using a single word or phrase."}
\begin{itemize}[leftmargin=*, labelsep=0.5em, itemsep=0pt]
    \item \textbf{Ground Truth Answer}: "luggage"
    \item \textbf{Vanilla Model}: "backpack"
    \item \textbf{Ours}: "suitcase"
\end{itemize}

\subsection{"What is in front of the poster?\textbackslash nAnswer the question using a single word or phrase."}
\begin{itemize}[leftmargin=*, labelsep=0.5em, itemsep=0pt]
    \item \textbf{Ground Truth Answer}: "monitor"
    \item \textbf{Vanilla Model}: "monitor"
    \item \textbf{Ours}: "computer"
\end{itemize}

\end{document}